\documentclass{article}

\usepackage{arxiv}

\usepackage[utf8]{inputenc} 
\usepackage[T1]{fontenc}    
\usepackage{url}            
\usepackage{booktabs}       
\usepackage{amsfonts}       
\usepackage{nicefrac}       
\usepackage{microtype}      
\usepackage{lipsum}

\usepackage{algorithm}      
\usepackage[noend]{algpseudocode}  
\usepackage{array}
\usepackage{amssymb,amsmath}
\usepackage{graphicx}

\title{Analyzing Reinforcement Learning Benchmarks with Random Weight Guessing}

\author{
  Declan Oller \\
  Providence, \\
  Rhode Island, USA \\
  \texttt{declanoller@gmail.com} \\
 \AND
 Tobias Glasmachers \\
  Institute for Neural Computation, \\
  Ruhr-University Bochum, \\
  Germany \\
  \texttt{tobias.glasmachers@ini.rub.de} \\
  \And
 Giuseppe Cuccu \\
  eXascale Infolab, \\
  University of Fribourg, \\
  Switzerland \\
  \texttt{giuse@exascale.info} \\
}

\begin{document}

\maketitle

\begin{abstract}  
We propose a novel method for analyzing and visualizing the complexity of standard reinforcement learning (RL) benchmarks based on score distributions.
A large number of policy networks are generated by randomly guessing their parameters, and then evaluated on the benchmark task; the study of their aggregated results provide insights into the benchmark complexity.
Our method guarantees objectivity of evaluation by sidestepping learning altogether: the policy network parameters are generated using Random Weight Guessing (RWG), making our method agnostic to (i) the classic RL setup, (ii) any learning algorithm, and (iii) hyperparameter tuning.
We show that this approach isolates the environment complexity, highlights specific types of challenges, and provides a proper foundation for the statistical analysis of the task's difficulty.
We test our approach on a variety of classic control benchmarks from the OpenAI Gym, where we show that small untrained networks can provide a robust baseline for a variety of tasks.
The networks generated often show good performance even without gradual learning, incidentally highlighting the triviality of a few popular benchmarks.
\end{abstract}


\section{Introduction}
\label{section:introduction}
Reports on new Reinforcement Learning (RL) algorithms \cite{sutton1998introduction} are frequently presented with the scores achieved on standard benchmark environments such as the OpenAI Gym~\cite{openai_gym}. These environments are chosen for their solvability and widespread use, meaning that scores for a new algorithm can be compared to methods in the literature that use the same environments. In many cases however we lack a solid understanding of the actual challenges posed by such environments~\cite{cuccu2019playing}. This limits the comparison to statements about general performance, and it precludes insights about specific strengths and limitations of RL algorithms. As commonly found in optimization, an ideal RL benchmark suite should offer variable, discriminative difficulty over a representative set of challenges. 

In optimization there exist well-understood challenges such as non-convexity, non-smoothness, multi-modality, ill-conditioning, noise, high dimensions, and others. They are attributed to certain problems, hence defining categories, and some of them even serve as numerical (hence quantitative) measures of problem hardness.
This work takes a first, major step in building an analogous measure of difficulty for RL tasks, by designing and open sourcing a tool that allows analysis of existing (and future) benchmarks based on an external and unbiased measure of complexity. It is understood that our proposal is not the only possible measure. In order to capture the many facets of task difficulty in RL we resist the temptation to capture complexity in a single number. Instead, we leverage the \emph{performance distribution} to visualize rich information about RL tasks.

We are concerned with measuring \emph{task difficulty} in a way that is appropriate for RL, and in particular for direct policy search. Instead of the optimization-related challenges listed above, we measure hardness in terms of the probability mass in the weight space that corresponds to acceptable solutions. \emph{It is in this sense that we address a fundamental problem of RL research}. We are not concerned with how to address specific difficulties with actual learning algorithms, like deep representation learning, exploration strategies, off-policy learning, and the like.

\begin{figure*}[!ht]
    \centering
    \-
    \hfill
    \fbox{\includegraphics[width=0.22\textwidth]{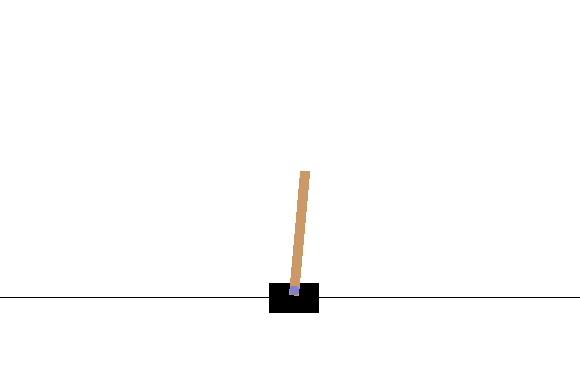}}
    \hfill
    \fbox{\includegraphics[trim=0 50 0 120, clip, width=0.22\textwidth]{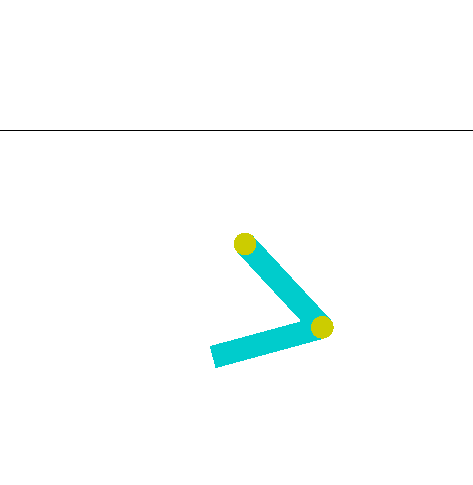}}
    \hfill
    \fbox{\includegraphics[trim=0 80 0 80, clip, width=0.22\textwidth]{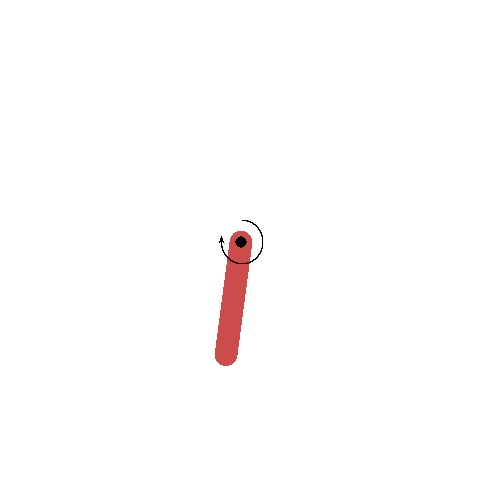}}
    \hfill
    \fbox{\includegraphics[width=0.22\textwidth]{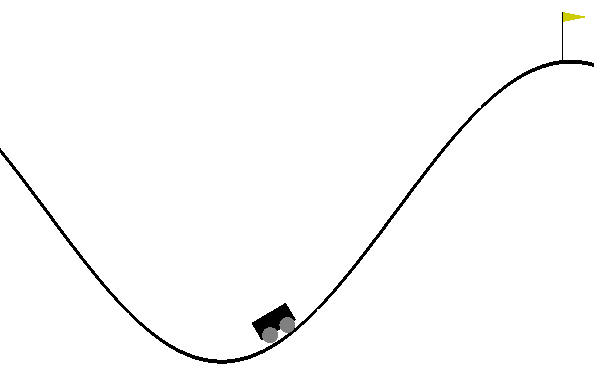}}
    \hfill
    \-
    \caption{
        Screenshots for the five Classic Control environments from the OpenAI Gym. From left to right: \texttt{CartPole}, \texttt{Acrobot} and \texttt{Pendulum}, while the last is representative of both \texttt{MountainCar} and \texttt{MountainCarContinuous}.
        \label{figure:environments}
    }
\end{figure*}

\begin{table*}[!ht]
\begin{center}
\begin{tabular}{m{35ex}rrrr}
        \toprule
        Environment name & \# observations & \# controls & control type & solved score \\
        \midrule
        \texttt{CartPole-v0} & 4 & 2 & discrete & 195 \\
        \texttt{Acrobot-v1} & 6 & 3 & discrete & -60 \\
        \texttt{Pendulum-v0} & 3 & 1 & continuous & -140 \\
        \texttt{MountainCar-v0} & 2 & 3 & discrete & -110 \\
        \texttt{MountainCarContinuous-v0} & 2 & 1 & continuous & 90 \\
        \bottomrule
\end{tabular}
\caption{Properties of the environments used in this study. The ``solved score'' is derived from current leader boards. Intuitively, performance close to or above this value corresponds to performance that is meaningful for the task.
}
\label{table:environments}
\end{center}
\end{table*}

Gauging a benchmark's complexity though is not a straightforward task.
As most RL environments are typically distributed as black-box software, the only way to interact with them that is guaranteed to be available is through the environment's control loop.
Exploring such interaction in a meaningful way requires a potentially sophisticated controller. It would be tempting to use any arbitrary learning approach for this task. Any choice of training method however would in turn unavoidably bias the analysis, limiting the extent and applicability of consequent findings.
The only viable alternative to maintain objectivity is thus to avoid learning completely.

For our analysis, we propose the simplest procedure conceivable: (i) direct policy search; (ii) controller models of minimal but increasing complexity; (iii) parameter selection based on a random sampling. 
In practice, this corresponds to selecting a Neural Network (NN) controller of minimal size and complexity, selecting its weights by means of Random Weight Guessing
(RWG;~\cite{schmidhuber2001evaluating})
, and directly using it in the observation-decision-action loop of control with the environment of choice.
There is no training, no gradient descent, no classical RL framework, and sampled controllers are completely independent from each other.
It's important to emphasize that RWG \emph{is not an approach for solving RL problems}, but rather an analysis method that can complement any learning strategy.
The extreme simplicity of RWG for parameter assignment presents several attractive benefits:
\begin{enumerate}
\item
    if obtaining controllers of satisfactory performance is likely, it directly raises the question of whether the environment constitutes a useful, nontrivial benchmark for more powerful algorithms;
\item
    it is easy to reproduce and hence especially useful as a baseline, as it does not depend on implementation details (such as software ecosystem) nor on meta-optimization techniques (such as hyperparameters tuning);
\item
    it does not suffer from random seed issues such as the methods investigated in~\cite{islam2017reproducibility}; and
\item
    it properly reflects the properties of the benchmarks, despite the curse of dimensionality, i.e., the high volume of the weight space it involves searching.
\end{enumerate}

On the topic of hyperparameters, RWG performance is dependant on the neural network architecture of choice, and on the distribution from which weights are drawn. We propose addressing the first point by performing a thorough parameter study, where a set of architectures of increasing model complexity is tested in turn. The choice of the distribution instead turns out to be uncritical: our results are nearly invariant under the shape of the distribution (we tested uniform and Gaussian), and unit variance turns out to be a solid default.
One should keep in mind that the bounds of the distribution (be there hard bounds as in the case of uniform distributions, or soft bounds as with the variance of a Gaussian) define the tested value range for the weights, which may be critical depending on application and network size (e.g.\ with the number of weights entering a neuron). However, in our experiments we did not find a need to tune the scale.

It is important to notice at this point that the weights generated by RWG are the final weights used, as opposed to initial configurations followed by learning, as is more common in deep learning.
For example, a concern often linked to the magnitude of the initial weights is neuron saturation: if the weighted sum of a neuron's input is high enough, the output of the sigmoidal activation function will be asymptotically close to its bounds, making its behavior indistinguishable from a step function.
This in turn squashes the error gradient to zero, hindering learning.
As described however in our case we are not interested in further learning: generated networks (with or without neuron saturation) are feasible solutions, and the overall performance of the network will be captured and interpreted by the final score on the task.

While neuron saturation limits the information capacity of the network~\cite{rakitianskaia2015measuring}, there is in principle no correlation with lower network performance: the generated network is final and as such evaluated uniquely based on its performance on the task, not based on the potential of networks that can be derived from it through learning.
In any case, our settings make neuron saturation unlikely: while the value range for weight generation is orders of magnitudes higher than commonly used in initialization in deep learning, the networks proposed in this work are also orders of magnitude smaller: the number of connections entering each neuron (and hence the number of terms in the weighted sum) is, proportionally, orders of magnitude less, effectively balancing weight magnitude.

Our experimental results address a set of broadly adopted RL benchmarks available from the OpenAI Gym~\cite{openai_gym}, specifically the Classic Control suite of benchmark environments\footnote{\url{https://gym.openai.com/envs/\#classic_control}}. A framework that we open source takes a benchmark environment and studies it by interacting with it using a sequence of increasingly complex neural networks. The proposed procedure is immediately applicable to any RL benchmark implementing the widespread OpenAI Gym interface, including both discrete and continuous control tasks.

We find that the statistical study derived from the distributions of scores highlights large qualitative differences among the tested environments. The information made available provides far more detailed insights than standard aggregated performance scores commonly reported in the literature.

\begin{figure*}[!ht]
    \centering
    \includegraphics[width=\textwidth]{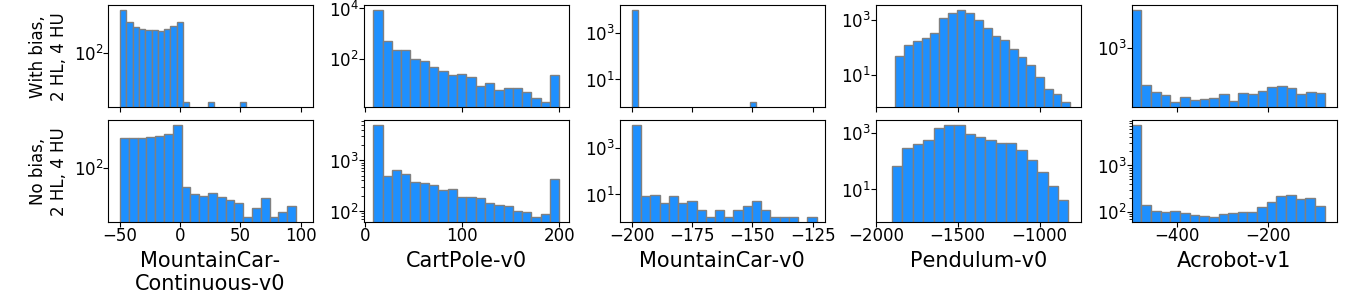}
    \caption{
        Histograms of mean sample scores for 2 hidden layer, 4 hidden unit networks \emph{with} (top row) and \emph{without} (bottom row) bias connections. In all five environments, the probability mass on top-performers generally increases when dropping bias connections. The difference is particularly striking for the  \texttt{MountainCarContinuous} and \texttt{MountainCar} environments.
    \label{figure:bias}
    }
\end{figure*}

\paragraph{Related Work}
RWG refers to applying the simplistic and usually non-competitive optimization technique of pure random search%
\footnote{Pure random search is a simplistic optimization technique drawing search points independently from a fixed probability distribution. Its search distribution is memory free, but the algorithm keeps track of the best point so far. It can be considered as a randomized variant of exhaustive search without taboo list.}
to the weights of neural networks. To the best of our knowledge it was first used in~\cite{schmidhuber2001evaluating} for demonstrating that certain widely used benchmark problems in sequence learning with recurrent neural networks are trivial. In contrast, we consider RL benchmarks, and our work develops the approach further by turning it into a useful analysis methodology for RL environments.

Pure random search was also proposed for tuning the hyperparameters of learning machines~\cite{bergstra2012random}, although it is usually less efficient than Bayesian optimization~\cite{jones1998efficient}. However, most Bayesian approaches start out with an initial design, which can be based on a random sample generated by RWG.

Random weight guessing by itself can be considered a simplistic baseline method for direct policy search learners such as Neuroevolution~\cite{igel2003neuroevolution}. In contrast to RWG, Evolution Strategies and similar algorithms adapt the search distribution online, which turns them into competitive RL methods~\cite{salimans2017evolution,muller2018challenges,cuccu2019playing}. In a direct policy search context, RWG can be thought of as a pure non-local exploration method.

It is crucial to distinguish RWG from such methods, since they sometimes come ``in 	disguise'', e.g., using the term \emph{random search} for elaborate procedures that do actually adapt distribution parameters and hence demonstrate competitive performance on non-trivial tasks \cite{mania2018simple}. We argue that such algorithms should instead be termed \emph{randomized search}, and they are best understood as zeroth order cousins of stochastic gradient descent (SGD). Although such algorithms can be simplistic (as is SGD) they clearly perform iterative learning, and hence their performance can hardly be considered unbiased measures of task difficulty.

Understanding the overall complexity of an RL task is usually approached through theoretical analysis, which yields general results, e.g., in terms of regret bounds~\cite{jaksch2010near} and convergence guarantees~\cite{sutton2018reinforcement,liu2015finite,silver2014deterministic,schulman2015trust}.
That line of work highlights principal strengths and limitations of RL paradigms, but it is hardly suitable for analyzing and systematically understanding specific challenges posed by complex RL environments. In addition, theoretically promised results of RL often break down when parts of the so-called ``Deadly Triad''%
\footnote{The term refers to combining function approximation, bootstrapping, and off-policy learning.}
are violated \cite{van2018deep}.
In particular, we are not aware of any work that considers widely used benchmarks like the OpenAI Gym collection of tasks in this perspective. Our study aims to provide a practical tool for such an analysis.

\paragraph{Our Contributions}
The above lines of work are mostly concerned with learning and optimization, as well as with myth busting~\cite{schmidhuber2001evaluating}. Here we propose to use RWG for a different purpose, namely for the analysis of RL tasks. Our contributions are as follows:
\begin{itemize}
\item
    We show that RWG is a surprisingly powerful method for the analysis of RL environments. In particular, the performance distribution provides information about different solution regimes, noise, and the minimal sophistication of a NN that can act as an effective controller.
\item
	We provide compact and informative visualizations of task difficulty that help to identify specific challenges and differences between environments at a single glance.
\item
	Nearly 20 years after the work of Schmidhuber et al. ~\cite{schmidhuber2001evaluating} we show that it is still common to (often unwittingly) report state-of-the-art results on benchmarks which should rather be considered trivial.%
	\footnote{\url{https://github.com/openai/gym/wiki/Leaderboard}}
\item
	We provide an open source software framework\footnote{\url{https://github.com/declanoller/rwg-benchmark}} to the community that allows immediate application of our analysis to any environment implementing the OpenAI Gym interface.
\end{itemize}

The remainder of this paper is organized as follows. In the next section we describe in detail how we turn RWG into a methodology for the analysis of RL environments. The following section demonstrates its discrimination and explanation power with exemplary results on selected OpenAI Gym environments. We close with our conclusions.


\section{Analysis Methodology}
\label{section:method}

In this section we describe our methodology. Given an RL environment we conduct a fixed series of experiments as follows.

\paragraph{Neural Network Controllers}
We construct a series of NN architectures suitable for the task at hand with the following implementation:
\begin{itemize}
\item
    The NN input and output sizes match the dimensionality of the environment observation and action spaces, respectively. These numbers are specific to the environment of choice.
\item
    The number of hidden layers and their sizes are specified as an experiment parameter (discussed below).
\item
    All networks use a non-linear \texttt{tanh} activation function.
\item
    In environments with a discrete action space, the output is translated into an action index with the $\arg\max$ operator.
\item
    For continuous action spaces, each output is scaled to within the action boundaries.
\item
    We consider $N_\text{architectures} = 3$ simple connectivity patterns of increasing complexity:
    \begin{itemize}
        \item The simplest case of a network without hidden layers, which is also equivalent to a linear model
        \item A network with a single hidden layer of 4 units
        \item A network with two hidden layers of 4 units each
    \end{itemize}

    Alternative architectures were tried without discernible additional insight, though of course benchmarks of higher complexity may benefit from larger controller models.
\item 
    All networks considered are tested  with and without bias inputs to the neurons, which interestingly produces better performance in the absence of bias (see Figure~\ref{figure:bias}).
\end{itemize}

For each of the network architectures we sample $N_\text{samples} = 10^4$ instances by drawing each of their weights i.i.d.\ from the standard normal distribution $\mathcal{N}(0, 1)$,
i.e.\ we draw $N_\text{samples}$ weight vectors $w_n \in \mathbb{R}^d$ from the multi-variate standard normal distribution $\mathcal{N}(0, I)$, where $d$ is the number of weights of a network architecture, and $I \in \mathbb{R}^{d \times d}$ denotes the identity matrix. Equivalently, a random matrix of size $N_\text{samples} \times d$ is drawn from the corresponding multi-variate standard normal distribution.

\begin{figure*}[!ht]
    \centering
    \includegraphics[width=0.33\textwidth]{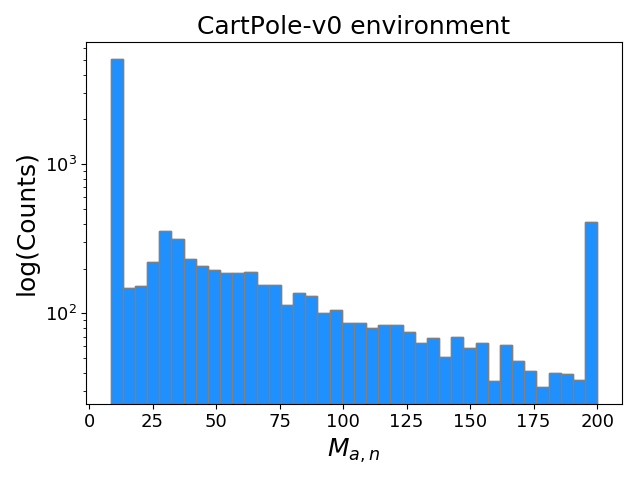}
    \includegraphics[width=0.33\textwidth]{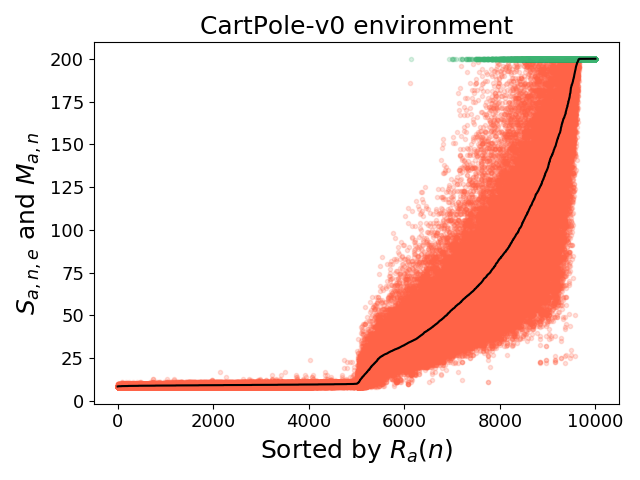}
    \includegraphics[width=0.33\textwidth]{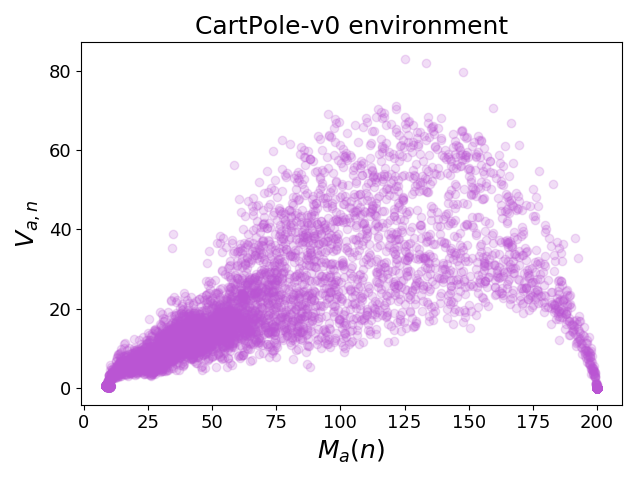}
    \caption{
        Plots of aggregate statistics produced by RWG, here for a network without hidden layers on the \texttt{CartPole} environment. From left to right: histogram of mean scores (note the log-scale), scatter plot of scores over rank sorted by mean score, and scatter plot of score variance over mean score. In the left plot, the counts decrease with mean episode score until the sharp increase of the highest score bin (scores 195 - 200), indicating that in general higher scores are harder to achieve, aside from a non-negligible number of samples that fully solve the problem. In the center plot, the top performing $0.1\%$ of all episodes (from all sampled networks) is colored in green.
        \label{figure:example-plots}
    }
\end{figure*}

\paragraph{The Score Tensor}
Each of the $N_\text{samples}$ networks implements a controller, which maps observations (input) to actions (output). Each controller is repeatedly tested on the environment's control loop for $N_\text{episodes}$ independent episodes. Repetition is only necessary for environments featuring a stochastic component, as common with random initial conditions. For deterministic environments we propose setting $N_\text{episodes} = 1$.

In each episode we record the rewards obtained in each time step. The total reward (the non-discounted sum of all rewards in the episode) is assigned as a score. For clarity, the procedure is formally defined in Algorithm~\ref{alg:RWG_alg1}.

The resulting score tensor has dimensions $N_\text{architectures} \times N_\text{samples} \times N_\text{episodes}$, for $N_\text{architectures} = 3$ network architectures, $N_\text{samples} = 10^4$ independent networks per architecture, and $N_\text{episodes} = 20$ independent episodes per network. We denote the tensor by $S$, and refer to the score achieved by network $n$ of architecture $a$ in the $e$-th episode as $S_{a,n,e}$.

\begin{algorithm}
\caption{Environment evaluation}
\label{alg:RWG_alg1}
\begin{algorithmic}
\State Initialize environment
\State Create array $S$ of size $N_\text{architectures} \times N_\text{samples} \times N_\text{episodes}$
\For{$a = 1, 2, \ldots, N_\text{architectures}$}
    \State Initialize the current NN architecture
    \For{$n = 1, 2, \ldots, N_\text{samples}$}
        \State Sample NN weights randomly from $\mathcal{N}(0, 1)$
        \For{$e = 1, 2, \ldots, N_\text{episodes}$}
            \State Reset the environment
            \State Run episode with NN
            \State Store accrued episode reward in $S_{a,n,e}$
        \EndFor
    \EndFor
\EndFor

\end{algorithmic}
\end{algorithm}

\begin{table*}[!ht]
\begin{center}
\begin{tabular}{l r r r}
        \toprule
        Environment & 0 HL & 1 HL, 4 HU & 2 HL, 4 HU \\
        \midrule
        \texttt{CartPole-v0} & 189.7 & 210.1 & 214.1 \\
        \texttt{Pendulum-v0} & 3206.8 & 3677.6 & 3438.3 \\
        \texttt{MountainCar-v0} & 2969.4 & 2808.4 & 2718.7 \\
        \texttt{MountainCarContinuous-v0} & 1820.9 & 1928.6 & 2305.8 \\
        \texttt{Acrobot-v1} & 14368.8 & 14611.0 & 14956.1 \\
        \bottomrule
\end{tabular}
\caption{Runtime (seconds) for the 3 architectures and 5 environments investigated. Statistics for each pair were collected for $10^4$ samples.}
\label{table:runtimes}
\end{center}
\end{table*}

A benefit of this search algorithm's simplicity and the independence between network architectures, networks, and episodes is that the algorithm is embarrassingly parallel. Also, our default numbers for $N_\text{samples}$ and $N_\text{episodes}$ can be adapted based on the available computational resources.
Running the full evaluation as described above on all five Classic Control environments from the OpenAI Gym (Figure~\ref{figure:environments})
with three architectures took under 20 hours on a single machine, using a 32-core Intel(R) Xeon(R) E5-2620 at 2.10GHz, with a RAM utilization below 5GB. Runtimes per architecture and environment are shown in Table \ref{table:runtimes}.

It is understood in principle that generating $N_\text{samples} = 10^4$ random networks is not a reasonable learning strategy. The number is far too large, e.g., for an initial design of a Bayesian optimization approach, or as a population size in a neuroevolution algorithm. We want to emphasize that such a large set of samples here serves a very different purpose: we do not aim to optimize the score this way, but instead we aim to draw statistical conclusions about properties of the environment.

\paragraph{The Score Distribution}
We visualize the score tensor $S$ as follows. First of all, each network architecture $a$ is treated independently, resulting in series of plots. For each network architecture the scores $S_a$ form an $N_\text{samples} \times N_\text{episodes}$ matrix. From each row of this matrix, corresponding to a single network $n$, we extract the mean performance
$$M_{a,n} = \frac{1}{N_\text{episodes}} \sum_{e=1}^{N_\text{episodes}} S_{a,n,e}$$
of the network, as well as its variance
$$V_{a,n} = \frac{1}{N_\text{episodes} + 1} \sum_{e=1}^{N_\text{episodes}} \Big(S_{a,n,e} - M_{a,n}\Big)^2.$$
A crucial ingredient of the subsequent analysis is that we sort the all networks of the same architecture by their mean score $M_{a,n}$, and we refer to the position of the network within the sorted list as its rank $R_a(n)$ (in the rare cases of exact ties of mean scores, the tied samples are left in the order they were originally). Then we aggregate the score matrix in three distinct figures:
\begin{enumerate}
\item
    A log-scale histogram of $M_{a,n}$;
\item
    A scatter plot of the individual sample scores $S_{a,n,e}$ over their corresponding $R_a(n)$ sorted by mean score, with a (naturally monotonically increasing) curve of their $M_{a,n}$ overlaid; and
\item
    A scatter plot of score variance $V_{a,n}$ over mean score $M_{a,n}$.
\end{enumerate}

Figure~\ref{figure:example-plots} shows an example visualization for a single-layer network applied to the OpenAI Gym \texttt{CartPole} environment. A number of interesting properties can be extracted:
\begin{itemize}
\item
    \textbf{Left.} Unsurprisingly, the vast majority of networks achieve relatively low scores. The histogram of mean scores decreasing throughout most of its range (mean scores 10-190) reflects our intuition that achieving higher scores is increasingly difficult. 
    However, the fraction of networks solving the task is surprisingly high, with over $3\%$ of the randomly generated networks achieving an average score sufficient to solve the task (above 195; see Table \ref{table:environments}). We can conclude that the task is trivial, and probably not a suitable benchmark for testing sophisticated RL algorithms.
\item
    \textbf{Center.} The distribution of the mean scores, indicated by the black curve, shows a smooth slope without jumps. This indicates that many RL algorithms should find it easy to learn the task incrementally with low risk of converging to local optima. In contrast, significant jumps in the curve would indicate the existence of distinct controller regimes, suggesting the need for algorithms capable of disruptive exploration (such as restart strategies). The highlighted green points at the top are the scores of the highest $0.1\%$ scores of all episodes (across all sampled controllers). The locations of these points indicate that a number of controllers with relatively low mean scores are often able to solve the environment, depending on initialization conditions. The right algorithm could exploit such early partial successes to significantly influence learning trajectories.
\item
    \textbf{Right}. The variance at lowest mean score is minimal, corresponding to controllers incapable of succeeding in the task regardless of initial conditions. In the mid-plot, scores instead have a wide spread, indicating controllers that achieve their mean score by succeeding on only some episodes and failing at others. For a controller to reach the highest mean score, consistency is required, as indicated by the tight variance at the rightmost end of the plot. 
    Overall the plot highlights a significant noise in the non-deterministic fitness evaluation, which may conceivably pose a barrier to some learning algorithms. At the same time other methods may instead thrive by focusing on the successful episodes, highlighting the importance of such a study for proper method selection. It is important to note that this specific behavior is not general to all possible environments; while it is conceivable that in some environments the optimal strategy could yield substantial variance, for all tasks considered in this study optimality goes hand in hand with stability.
\end{itemize}

\begin{figure*}[!h]
    \centering
    \vspace*{5ex}
    \includegraphics[width=\textwidth]{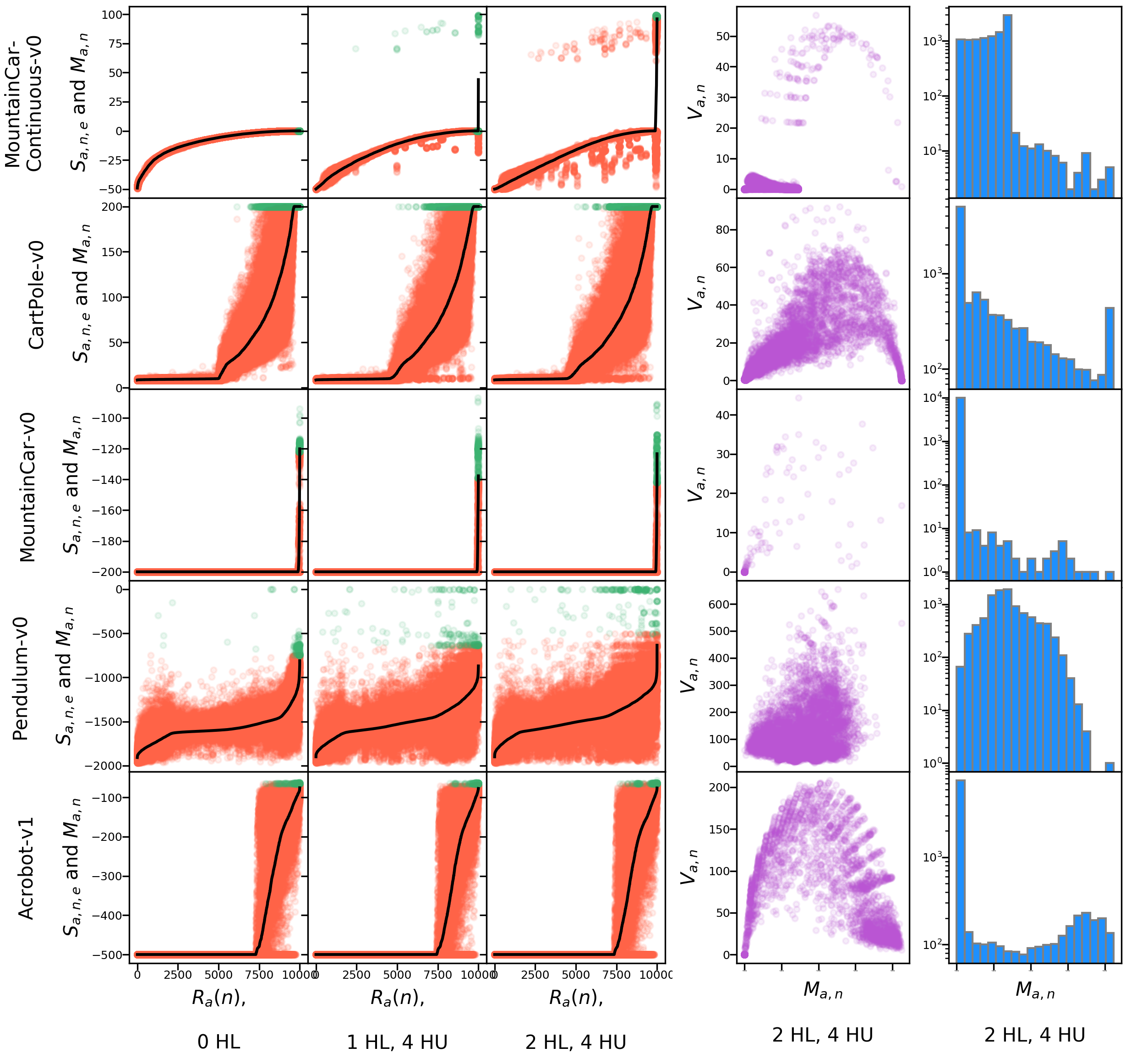}
    \vspace*{3ex}
    \caption{
        Score distribution plots for all five classic control tasks. Each row of five plots corresponds to one environment. The first three columns show scatter plots of episode scores for the three network architectures studied. The green dots in the scatter plots represent the $0.1\%$ best episodes. The fourth column shows a scatter plot of score variance over mean score, and the fifth column shows the score histogram, both for networks with 2 hidden layers of 4 units each (2 HL, 4 HU). Plots in the same row of the fourth and fifth columns share the same x-axis range, which has been scaled to the full width for clarity. The exact ranges can be read in figure \ref{figure:bias}.
        \label{figure:grid}
    }
\end{figure*}


\begin{table*}[!ht]
\begin{center}
\begin{tabular}{l r r c r r c r r }
\toprule
Environment & \multicolumn{2}{r}{0 HL} && \multicolumn{2}{r}{1 HL, 4 HU} && \multicolumn{2}{r}{2 HL, 4 HU} \\
\midrule
\texttt{CartPole-v0} & 200.0 &(200.0) && 200.0 &(200.0) && 200.0 &(200.0) \\
\texttt{Pendulum-v0} & -811.6 &(-916.7) && -870.4 &(-932.5) && -635.3 &(-939.5) \\
\texttt{MountainCar-v0} & -120.2 &(-144.2) && -137.7 &(-147.6) && -123.3 &(-149.2) \\
\texttt{MountainCarContinuous-v0} & -0.0 &(-0.0) && 44.0 &(4.1) && 96.1 &(73.5) \\
\texttt{Acrobot-v1} & -75.6 &(-80.5) && -77.3 &(-80.2) && -76.0 &(-81.4) \\
\bottomrule
\end{tabular}
\caption{Best mean sample score found and 99.9th percentile score (in parentheses) for each architecture and environment combination.}
\label{table:perc_99_scores_table_no_bias}
\end{center}
\end{table*}
\begin{table*}[!ht]
\begin{center}
\begin{tabular}{l r r c r r c r r }
\toprule
Environment & \multicolumn{2}{r}{0 HL} && \multicolumn{2}{r}{1 HL, 4 HU} && \multicolumn{2}{r}{2 HL, 4 HU} \\
\midrule
\texttt{CartPole-v0} & 200.0 &(200.0) && 200.0 &(200.0) && 200.0 &(200.0) \\
\texttt{Pendulum-v0} & -828.2 &(-1052.5) && -797.0 &(-980.1) && -813.8 &(-969.3) \\
\texttt{MountainCar-v0} & -158.1 &(-200.0) && -136.3 &(-200.0) && -139.4 &(-200.0) \\
\texttt{MountainCarContinuous-v0} & -0.0 &(-0.0) && 38.1 &(-0.0) && 74.2 &(-0.0) \\
\texttt{Acrobot-v1} & -72.8 &(-79.9) && -73.8 &(-81.4) && -73.8 &(-81.5) \\
\bottomrule
\end{tabular}
\caption{Best mean sample score found and 99.9th percentile score (in parentheses) for each architecture and environment combination, for networks with bias connections. Comparison with table \ref{table:perc_99_scores_table_no_bias} shows that for the majority of environment and architecture pairs, having a bias node hurts performance.}
\label{table:perc_99_scores_table_with_bias}
\end{center}
\end{table*}


\section{Example Study: OpenAI Classic Control}
\label{section:experiments}

In this section we demonstrate that significant insights about environments and their differences can be gained for the procedure above. To this end we consider the five ``classic control'' environments from the OpenAI Gym collection (Table~\ref{table:environments}).
Performance on Gym environments is typically measured in one of two ways:
\begin{enumerate}
\item
    by defining a threshold ``solved score'', and reporting the number of training/search episodes it takes until a given algorithm can achieve this score as its mean score across 100 consecutive episodes, or
\item
    by reporting the best score the controller achieves on average across 100 consecutive episodes.
\end{enumerate}
Both methods are imperfect: method 1) does not account for the (often significant) number of episodes evaluated during hyperparameter search, and can also depend critically on random effects such as experiencing an early reward in sparse environments; method 2) provides an upper bound for the performance, but does not take computation and learning time into account.

Our analysis proposes a more sensible, objective aggregation for the run results. It is easy to read off an estimate of the probability $p$ of achieving a certain score $\geq s$ (e.g., the solved score), and from that number we can trivially derive the probability $1 - (1-p)^N$ of solving an environment at least once with a given number $N$ of sampled networks. The waiting time for this event follows a geometric distribution with parameter~$p$, hence the expected waiting time in simply $1/p$, which represents the expected run time of RWG until hitting that score.

\paragraph{Results}
We ran the analysis procedure described in the previous section on the tasks listed in Table~\ref{table:environments}.
Selected representative results are listed in Table~\ref{table:perc_99_scores_table_no_bias} and Figure~\ref{figure:grid}. Figure~\ref{figure:bias} demonstrates our surprising findings on NN performance with and without bias.

\paragraph{Discussion}
The following effects are observed:
\begin{itemize}
\item
    Figure~\ref{figure:grid} highlights the environments showing qualitatively very different characteristics, while the network architecture has a minor impact on the overall shape of the plots. This is due to the vast majority of NN samples of any architecture having relatively poor performance.
    A closer inspection of the top performers however is available in Table \ref{table:perc_99_scores_table_no_bias}, with larger networks performing distinctively better for some environments.
\item
    Some environments (e.g.\ \texttt{CartPole}, \texttt{MountainCar}, \newline \texttt{MountainCarContinuous}) can be solved with rather simple networks, or even with linear controllers. In some cases the probability of randomly guessing a successful controller is so high that the benchmark can be considered trivial, as it could be solved by straight RWG in reasonable time. This is clearly the case for the \texttt{CartPole} environment.
\item
    The score histograms and mean score curves in Figure~\ref{figure:grid} differ significantly across environments. This highlights the diversity of challenges posed for RL algorithms, in itself a desirable property for a benchmark suite.
\item
    With the exception of the \texttt{MountainCar} environment, all mean curves are nicely sloped (above a reasonably easy to find rank) and continuous, which means that gradual (e.g., gradient-based) iterative learning algorithms should be suitable in principle. For \texttt{MountainCar} these methods are instead confronted with a large plateau of minimal scores, which means that the task would be better addressed with methods strong in exploration.
\item
    The distribution of variance discussed in the last section, with low variance in correspondence of both the lowest and highest mean scores, is to be expected as a common pattern. All samples should conceivably fall into one of the following three categories: 1) never receives reward; 2) reward depends on favorable initial condition, but is inconsistent; and 3) reliably achieving good scores, independent of initial conditions. Categories 1 and 3 naturally lead to low variance, as the scores are either all low, or all high.

    This implies that the score variance is to be expected highest in the range where learning takes place. For the \texttt{CartPole}, \texttt{Acrobot}, and \texttt{Pendulum} tasks the variability covers a huge range of scores, which makes it difficult to even measure progress online during learning. We can therefore expect that despite the fact that learning trajectories with smoothly increasing mean scores exist, some types of RL algorithms (like direct policy search) may be expected to suffer significantly from the high score variance. In contrast, the \texttt{MountainCarContinuous} task is nearly unaffected by noise. In the interesting range it still has a large variance due to the fact that only few controllers solve the task in a few episodes, while the vast majority is caught in a large local optimum.
\item
    It is at first glance surprising that NN controllers without bias terms outperform NNs with bias (Figure~\ref{figure:bias}). We find that this effect is systematic across all tested environments, barring negligible statistical fluctuations. The analogous overview of results presented in Table \ref{table:perc_99_scores_table_no_bias}, but instead for NNs with bias units, is shown in Table \ref{table:perc_99_scores_table_with_bias}. 
\end{itemize}

Our open source reference implementation (in Python) makes it easy to extend the study to other classes of environments, especially if already compatible with the widely adopted OpenAI Gym control interface. Testing a large number of environments over time would create a broad data base of characterized environments, constituting a strong baseline for the study of existing and new learning methods.


\section{Conclusion}
\label{section:conclusion}

Evaluating task complexity in the context of reinforcement learning problems is a multifaceted and understudied problem. Rather than aiming at providing a single score of overall complexity, which would inevitably remain incomplete, we present a framework to analyze in depth the complexity of RL tasks.
Our analysis uses no learning and does not require any hyperparameter tuning. We produce test controllers in a direct policy search fashion using Random Weight Guessing, then draw a statistical analysis based on the complexity of the controllers, their performance on the task, and the distribution of collected reward.

We validate our approach on the set of Classic Control benchmarks from the OpenAI Gym.
Due to this limitation of the scope of our study we consider it only a first step. We nevertheless regard this step as an important contribution to a study subject that deserves more attention in the future.
Our results clearly identify the distinctive characteristics of each environment, underlying the challenges that induce their complexity, and pointing at promising approaches to address them. Moreover we offer an upper bound on required model complexity. We find RWG to be surprisingly effective e.g.\ in the case of the \texttt{CartPole} problem, pointing at its triviality.

\paragraph{Future Work}

One apparent limitation of RWG regards scaling to large network architectures, which (at first glance) seems to preclude its application to tasks relying on visual input such as Atari games~\cite{mnih2015human}.
In future work we will address this widely used class of benchmarks by separating feature extraction from the actual controller, following~\cite{cuccu2019playing}.
Yet another straightforward extension is to include recurrent NN architectures to better cope with partially observable environments.

A striking open question is how well our analysis predicts the performance of different classes of algorithms, like temporal difference approaches, policy gradient methods, and direct policy search. Answering this question would have the potential to extend our analysis methodology into a veritable recommender system for RL algorithms.

\section*{Acknowledgements}
This work was supported by the Swiss National Science Foundation under grant number 407540\_167320.

\bibliographystyle{unsrt}  


\begin{thebibliography}{10}

\bibitem{sutton1998introduction}
Richard~S Sutton, Andrew~G Barto, et~al.
\newblock {\em Introduction to reinforcement learning}, volume~2.
\newblock MIT press Cambridge, 1998.

\bibitem{openai_gym}
Greg Brockman, Vicki Cheung, Ludwig Pettersson, Jonas Schneider, John Schulman,
  Jie Tang, and Wojciech Zaremba.
\newblock {OpenAI Gym}, 2016.

\bibitem{cuccu2019playing}
Giuseppe Cuccu, Julian Togelius, and Philippe Cudr{\'e}-Mauroux.
\newblock Playing {A}tari with six neurons.
\newblock In {\em Proceedings of the 18th International Conference on
  Autonomous Agents and MultiAgent Systems}, pages 998--1006, 2019.

\bibitem{schmidhuber2001evaluating}
J{\"u}rgen Schmidhuber, S~Hochreiter, and Y~Bengio.
\newblock Evaluating benchmark problems by random guessing.
\newblock {\em A Field Guide to Dynamical Recurrent Networks, ed. J. Kolen and
  S. Cremer}, pages 231--235, 2001.

\bibitem{islam2017reproducibility}
Riashat Islam, Peter Henderson, Maziar Gomrokchi, and Doina Precup.
\newblock Reproducibility of benchmarked deep reinforcement learning tasks for
  continuous control.
\newblock {\em arXiv preprint arXiv:1708.04133}, 2017.

\bibitem{rakitianskaia2015measuring}
Anna Rakitianskaia and Andries Engelbrecht.
\newblock Measuring saturation in neural networks.
\newblock In {\em 2015 IEEE Symposium Series on Computational Intelligence},
  pages 1423--1430. IEEE, 2015.

\bibitem{bergstra2012random}
James Bergstra and Yoshua Bengio.
\newblock Random search for hyper-parameter optimization.
\newblock {\em Journal of Machine Learning Research (JMLR)}, 13:281--305, 2012.

\bibitem{jones1998efficient}
Donald~R Jones, Matthias Schonlau, and William~J Welch.
\newblock Efficient global optimization of expensive black-box functions.
\newblock {\em Journal of Global optimization}, 13(4):455--492, 1998.

\bibitem{igel2003neuroevolution}
Christian Igel.
\newblock Neuroevolution for reinforcement learning using evolution strategies.
\newblock In {\em The 2003 Congress on Evolutionary Computation, 2003.
  CEC'03.}, volume~4, pages 2588--2595. IEEE, 2003.

\bibitem{salimans2017evolution}
Tim Salimans, Jonathan Ho, Xi~Chen, Szymon Sidor, and Ilya Sutskever.
\newblock Evolution strategies as a scalable alternative to reinforcement
  learning.
\newblock Technical Report arXiv:1703.03864, arXiv.org, 2017.

\bibitem{muller2018challenges}
Nils M{\"u}ller and Tobias Glasmachers.
\newblock Challenges in high-dimensional reinforcement learning with evolution
  strategies.
\newblock In {\em International Conference on Parallel Problem Solving from
  Nature}, pages 411--423. Springer, 2018.

\bibitem{mania2018simple}
Horia Mania, Aurelia Guy, and Benjamin Recht.
\newblock Simple random search of static linear policies is competitive for
  reinforcement learning.
\newblock In {\em Neural Information Processing Systems}, volume~31, 2018.

\bibitem{jaksch2010near}
Thomas Jaksch, Ronald Ortner, and Peter Auer.
\newblock Near-optimal regret bounds for reinforcement learning.
\newblock {\em Journal of Machine Learning Research (JMLR)},
  11(Apr):1563--1600, 2010.

\bibitem{sutton2018reinforcement}
Richard~S Sutton and Andrew~G Barto.
\newblock {\em Reinforcement learning: An introduction}.
\newblock MIT press, 2018.

\bibitem{liu2015finite}
Bo~Liu, Ji~Liu, Mohammad Ghavamzadeh, Sridhar Mahadevan, and Marek Petrik.
\newblock Finite-sample analysis of proximal gradient td algorithms.
\newblock In {\em UAI}, pages 504--513, 2015.

\bibitem{silver2014deterministic}
David Silver, Guy Lever, Nicolas Heess, Thomas Degris, Daan Wierstra, and
  Martin Riedmiller.
\newblock Deterministic policy gradient algorithms.
\newblock In {\em JMLR conference proceedings, International Conference on
  Machine Learning}, volume~32, 2014.

\bibitem{schulman2015trust}
John Schulman, Sergey Levine, Pieter Abbeel, Michael Jordan, and Philipp
  Moritz.
\newblock Trust region policy optimization.
\newblock In {\em International conference on machine learning}, pages
  1889--1897, 2015.

\bibitem{van2018deep}
Hado Van~Hasselt, Yotam Doron, Florian Strub, Matteo Hessel, Nicolas Sonnerat,
  and Joseph Modayil.
\newblock Deep reinforcement learning and the deadly triad.
\newblock {\em arXiv preprint arXiv:1812.02648}, 2018.

\bibitem{mnih2015human}
Volodymyr Mnih, Koray Kavukcuoglu, David Silver, Andrei~A Rusu, Joel Veness,
  Marc~G Bellemare, Alex Graves, Martin Riedmiller, Andreas~K Fidjeland, Georg
  Ostrovski, et~al.
\newblock Human-level control through deep reinforcement learning.
\newblock {\em Nature}, 518(7540):529, 2015.

\end{thebibliography}

\newpage

\appendix
\section{Appendix}
\subsection{Full data plots for networks with no bias connections}

\begin{figure}[ht!]
    \centering
    \includegraphics[width=0.95\textwidth]{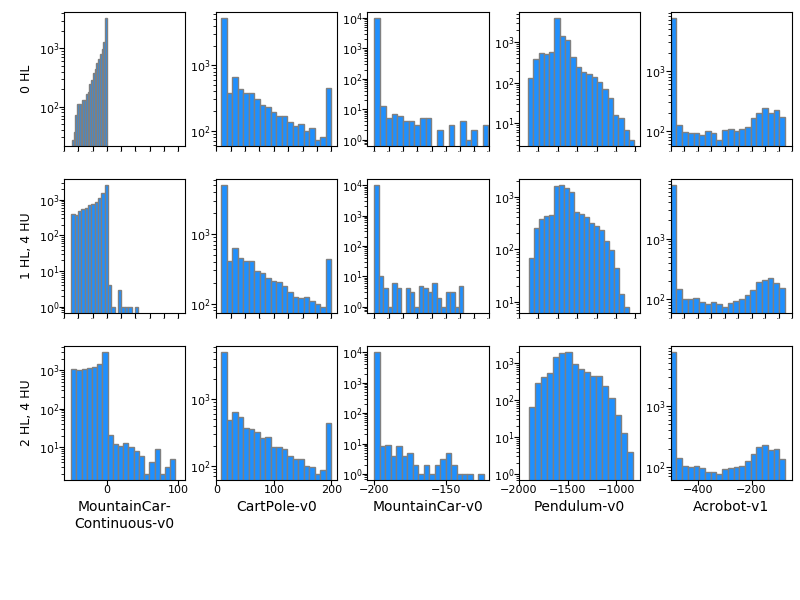}
    \caption{
    Log histograms for mean sample scores, as a function of NN architecture and environment, for NNs with no bias connections.
        \label{figure:supp_fig_1}
    }
\end{figure}

\begin{figure}[ht!]
    \centering
    \includegraphics[width=0.95\textwidth]{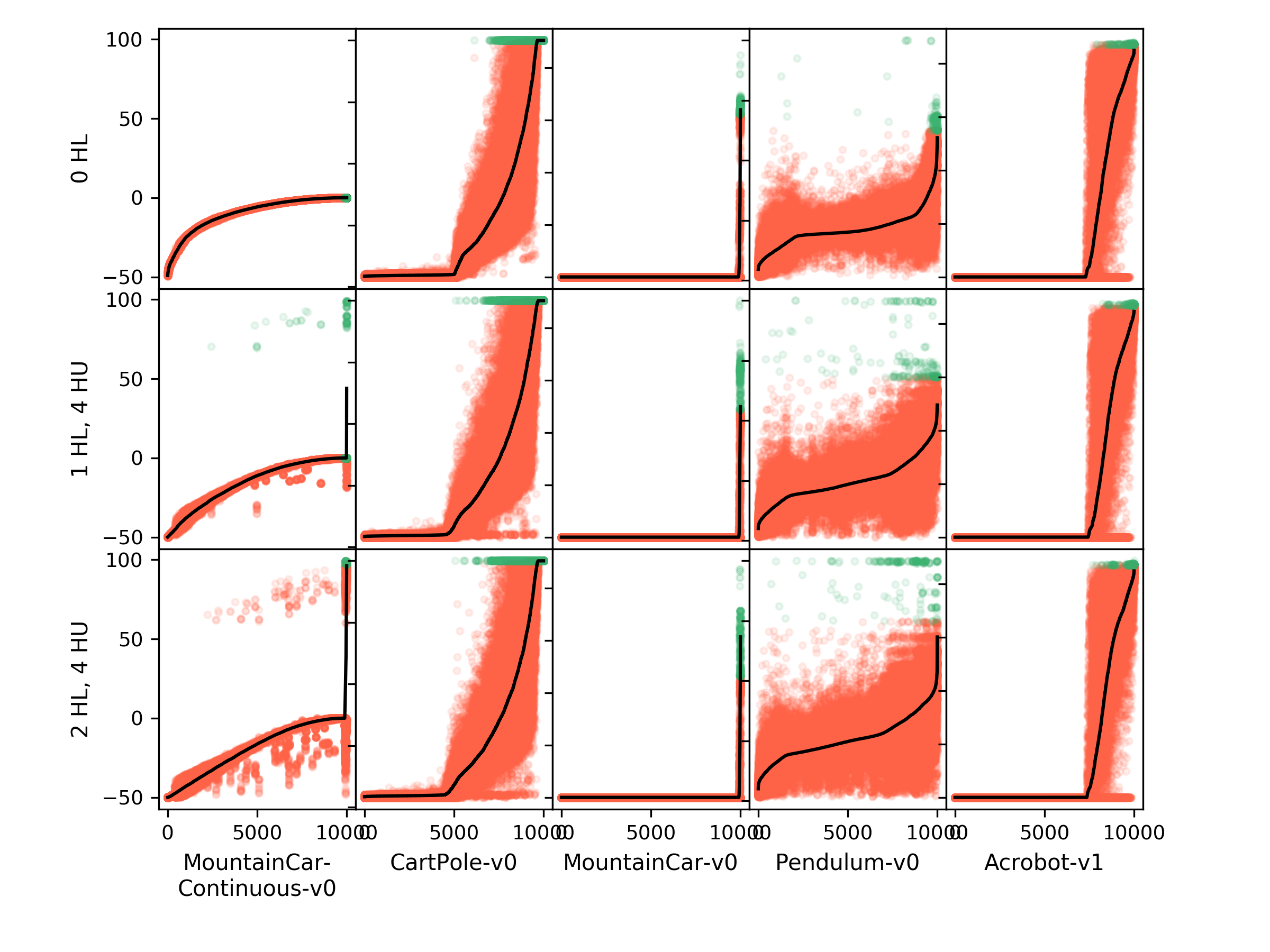}
    \caption{
    Scatter plots of all episode scores, as a function of NN architecture and environment, for NNs with no bias connections.
        \label{figure:supp_fig_2}
    }
\end{figure}

\begin{figure}[ht!]
    \centering
    \includegraphics[width=0.95\textwidth]{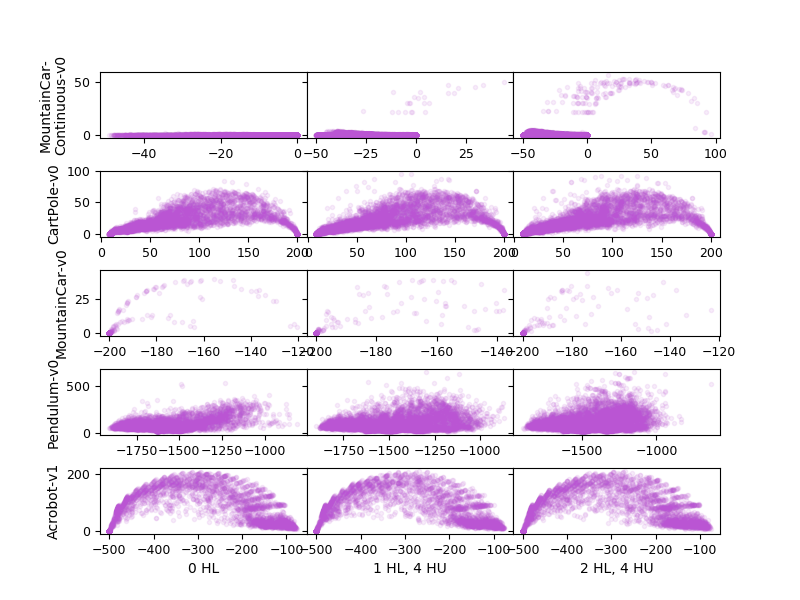}
    \caption{
    Scatter plots of sample score variances over mean sample scores, as a function of NN architecture and environment, for NNs with no bias connections.
        \label{figure:supp_fig_3}
    }
\end{figure}

\pagebreak

\subsection{Performance comparison with networks using bias connections}

\begin{figure}[ht!]
    \centering
    \includegraphics[width=0.95\textwidth]{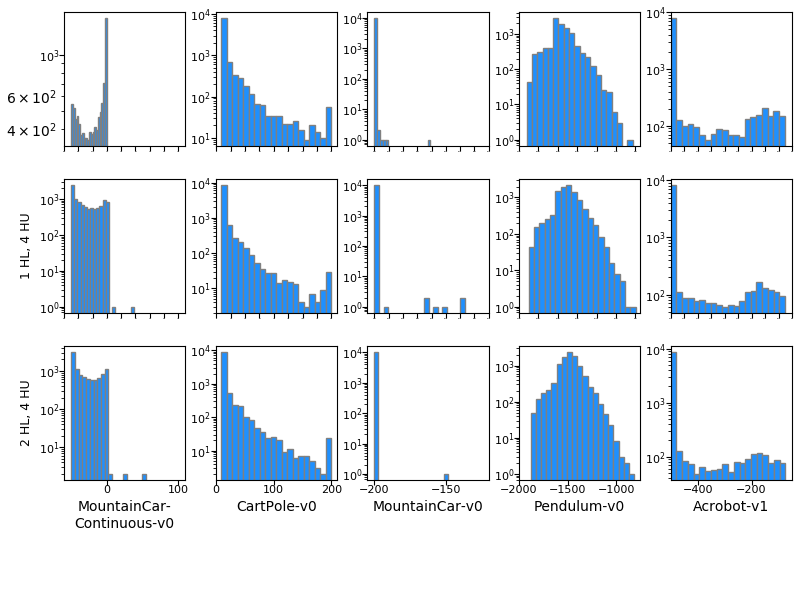}
    \caption{
    Log histograms for mean sample scores, as a function of NN architecture and environment, for NNs with bias connections.
        \label{figure:supp_fig_4}
    }
\end{figure}

\begin{figure}[ht!]
    \centering
    \includegraphics[width=0.95\textwidth]{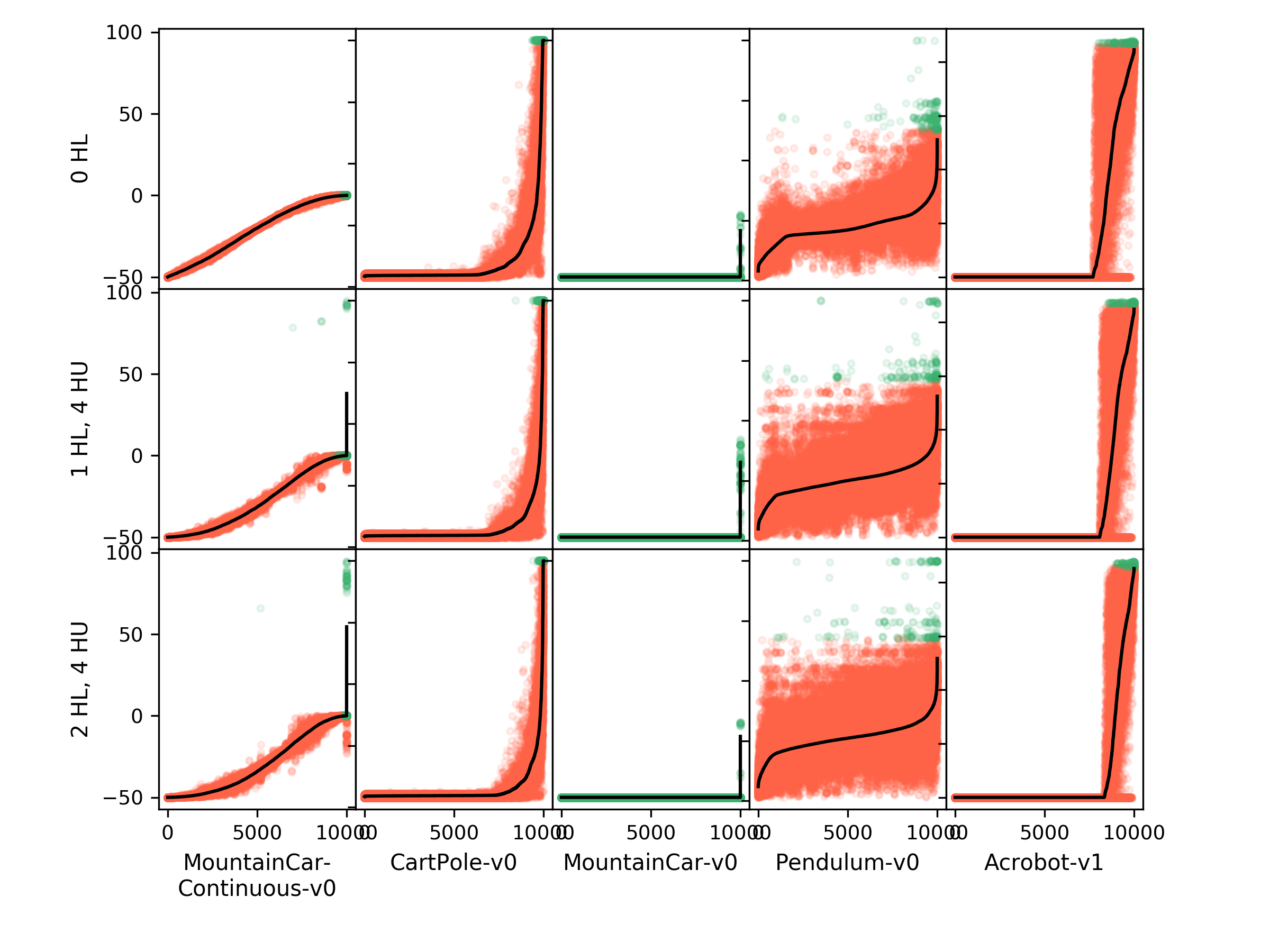}
    \caption{
    Scatter plots of all episode scores, as a function of NN architecture and environment, for NNs with bias connections.
        \label{figure:supp_fig_5}
    }
\end{figure}

\begin{figure}[ht!]
    \centering
    \includegraphics[width=\textwidth]{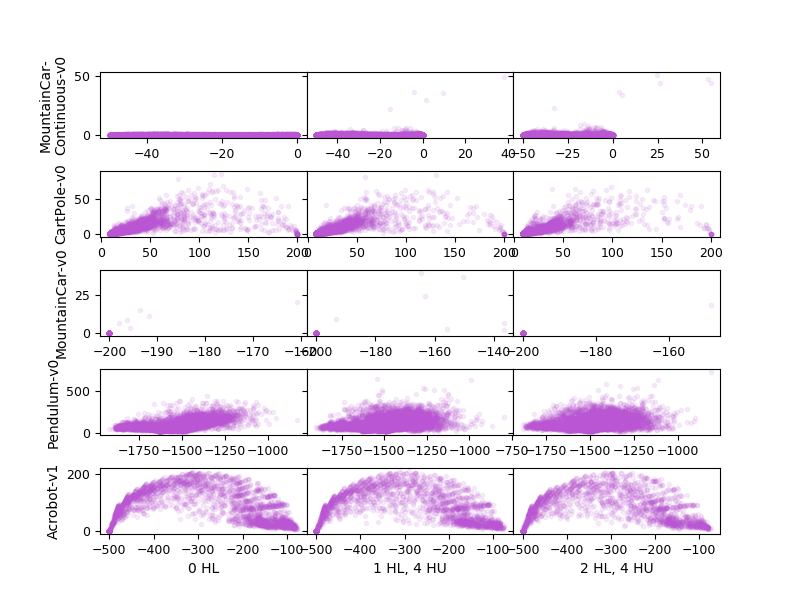}
    \caption{
    Scatter plots of sample score variances over mean sample scores, as a function of NN architecture and environment, for NNs with bias connections.
        \label{figure:supp_fig_6}
    }
\end{figure}

\begin{table}[h!]
\begin{center}
 \begin{tabular}{m{4.3cm} || m{2.4cm} | m{2.4cm} | m{2.4cm}}
\centering Environment & 0 HL & 1 HL, 4 HU & 2 HL, 4 HU \\
\hline\hline

\texttt{CartPole-v0} & 200.0 (200.0) & 200.0 (200.0) & 200.0 (200.0) \\
\hline
\texttt{Pendulum-v0} & -828.2 (-1052.5) & -797.0 (-980.1) & -813.8 (-969.3) \\
\hline
\texttt{MountainCar-v0} & -158.1 (-200.0) & -136.3 (-200.0) & -139.4 (-200.0) \\
\hline
\texttt{MountainCarContinuous-v0} & -0.0 (-0.0) & 38.1 (-0.0) & 74.2 (-0.0) \\
\hline
\texttt{Acrobot-v1} & -72.8 (-79.9) & -73.8 (-81.4) & -73.8 (-81.5) \\
\end{tabular}
\caption{Best mean sample score found and 99.9th percentile score (in parentheses) for each architecture and environment combination, for networks with bias connections.}
\label{table:supp_table_1}
\end{center}
\end{table}

\begin{figure}[ht!]
    \centering
    \includegraphics[width=\textwidth]{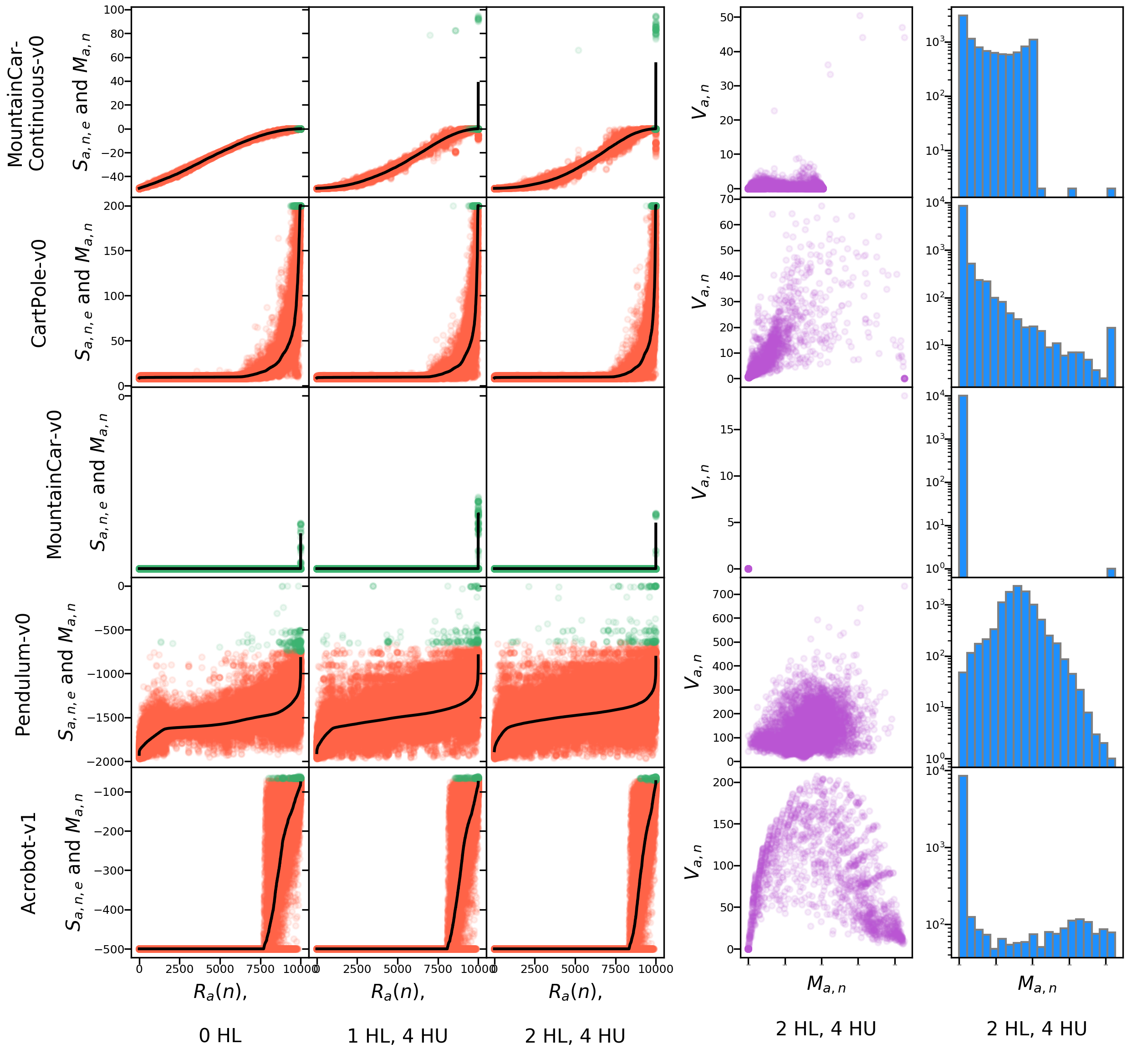}
    \caption{
        Score distribution plots for all five classic control tasks, with bias connections. Each row of five plots corresponds to one environment. The first three columns show scatter plots of episode scores for the three network architectures studied. The green dots in the scatter plots represent the $0.1\%$ best episodes. The fourth column shows a scatter plot of score variance over mean score, and the fifth column shows the score histogram, both for networks with 2 hidden layers of 4 units each (2 HL, 4 HU). Plots in the same row of the fourth and fifth columns share the same x-axis range, which has been scaled to the full width for clarity. The exact ranges can be read in figure 3 of the main text. 
        \label{figure:supp_fig_7}
    }
\end{figure}

\pagebreak

\subsection{$N_{weights}$ for different architectures and environments}

\subsubsection{No bias nodes}

\begin{table}[ht!]
\begin{center}
 \begin{tabular}{m{4.3cm} || m{0.6cm} | m{0.6cm} | m{0.8cm} | m{1.0cm} | m{1.0cm} | m{1.0cm} | m{1.0cm}}
\centering Environment & $N_{in}$ & $N_{out}$ & 0 HL & 1 HL, 2 HU & 1 HL, 4 HU & 1 HL, 8 HU & 2 HL, 4 HU \\
\hline\hline
\texttt{CartPole-v0} & 4 & 2 & 8 & 12 & 24 & 48 & 40 \\
        \hline
        \texttt{MountainCar-v0} & 2 & 3 & 6 & 10 & 20 & 40 & 36 \\
        \hline
        \texttt{MountainCarContinuous-v0} & 2 & 1 & 2 & 6 & 12 & 24 & 28 \\
        \hline
        \texttt{Pendulum-v0} & 3 & 1 & 3 & 8 & 16 & 32 & 32 \\
        \hline
        \texttt{Acrobot-v1} & 6 & 3 & 18 & 18 & 36 & 72 & 52 \\
\end{tabular}
\caption{Overview of the number of input and output nodes for each environment, as well as the number of weights each NN architecture has for that environment (no bias connections).}
\label{table:supp_table_2}
\end{center}
\end{table}

\subsubsection{With bias nodes}

\begin{table}[h!]
\begin{center}
 \begin{tabular}{m{4.3cm} || m{0.6cm} | m{0.6cm} | m{0.8cm} | m{1.0cm} | m{1.0cm} | m{1.0cm} | m{1.0cm}}
\centering Environment & $N_{in}$ & $N_{out}$ & 0 HL & 1 HL, 2 HU & 1 HL, 4 HU & 1 HL, 8 HU & 2 HL, 4 HU \\
\hline\hline
\texttt{CartPole-v0} & 4 & 2 & 10 & 16 & 30 & 58 & 50 \\
        \hline
        \texttt{MountainCar-v0} & 2 & 3 & 9 & 15 & 27 & 51 & 47 \\
        \hline
        \texttt{MountainCarContinuous-v0} & 2 & 1 & 3 & 9 & 17 & 33 & 37 \\
        \hline
        \texttt{Pendulum-v0} & 3 & 1 & 4 & 11 & 21 & 41 & 41 \\
        \hline
        \texttt{Acrobot-v1} & 6 & 3 & 21 & 23 & 43 & 83 & 63 \\
\end{tabular}

\caption{Overview of the number of input and output nodes for each environment, as well as the number of weights each NN architecture has for that environment (with bias connections).}
\label{table:supp_table_3}
\end{center}
\end{table}

\pagebreak

\subsection{Total runtime for different environments}

\begin{table}[h!]
\begin{center}
 \begin{tabular}{m{4.3cm} || m{1.0cm} | m{1.0cm} | m{1.0cm}}
\centering Environment & 0 HL & 1 HL, 4 HU & 2 HL, 4 HU \\
\hline\hline
\texttt{CartPole-v0} & 189.7 & 210.1 & 214.1 \\
        \hline
        \texttt{Pendulum-v0} & 3206.8 & 3677.6 & 3438.3 \\
        \hline
        \texttt{MountainCar-v0} & 2969.4 & 2808.4 & 2718.7 \\
        \hline
        \texttt{MountainCarContinuous-v0} & 1820.9 & 1928.6 & 2305.8 \\
        \hline
        \texttt{Acrobot-v1} & 14368.8 & 14611.0 & 14956.1 \\
\end{tabular}
\caption{Runtime (seconds) for 3 archs, 5 envs, with no bias connections.}
\label{table:supp_table_4}
\end{center}
\end{table}

\end{document}